\documentclass[conference]{IEEEtran}
\IEEEoverridecommandlockouts
\pdfoutput=1
\usepackage{cite}
\usepackage{amsmath,amssymb,amsfonts}
\usepackage{algorithmic}
\usepackage[table,xcdraw]{xcolor}  
\usepackage{graphicx}
\usepackage{textcomp}
\usepackage{xcolor}
\usepackage{float}
\usepackage{booktabs}
\usepackage{hyperref} 
\hypersetup{
    pdfborder={0 0 0},  
}
\usepackage{multirow}
\definecolor{mred}{RGB}{238, 34, 12}
\definecolor{mgreen}{RGB}{1, 127, 0}
\definecolor{mblue}{RGB}{0, 77, 158}

\def\BibTeX{{\rm B\kern-.05em{\sc i\kern-.025em b}\kern-.08em
    T\kern-.1667em\lower.7ex\hbox{E}\kern-.125emX}}

\begin{document}
\setlength{\topmargin}{-1in}
\setlength{\footskip}{-1in}
\title{HarmonyIQA: Pioneering Benchmark and Model for Image Harmonization Quality Assessment}

\author{Zitong Xu, Huiyu Duan, Guangji Ma, Liu Yang, Jiarui Wang, Qingbo Wu, \\Xiongkuo Min, Guangtao Zhai, Patrick Le Callet}

\maketitle

\begin{abstract}
Image composition involves extracting a foreground object from one image and pasting it into another image through Image harmonization algorithms (IHAs), which aim to adjust the appearance of the foreground object to better match the background. Existing image quality assessment (IQA) methods may fail to align with human visual preference on image harmonization due to the insensitivity to minor color or light inconsistency. To address the issue and facilitate the advancement of IHAs, we introduce the first Image Quality Assessment Database for image Harmony evaluation (HarmonyIQAD), which consists of 1,350 harmonized images generated by 9 different IHAs, and the corresponding human visual preference scores. Based on this database, we propose a Harmony Image Quality Assessment (HarmonyIQA), to predict human visual preference for harmonized images. Extensive experiments show that HarmonyIQA achieves state-of-the-art performance on human visual preference evaluation for harmonized images, and also achieves competing results on traditional IQA tasks. Furthermore, cross-dataset evaluation also shows that HarmonyIQA exhibits better generalization ability than self-supervised learning-based IQA methods. Both HarmonyIQAD and HarmonyIQA will be made publicly available upon paper publication.
\end{abstract}

\vspace{-8pt}
\section{Introduction}
\begin{figure}[tb]  
\centering  
\includegraphics[width=0.49\textwidth]{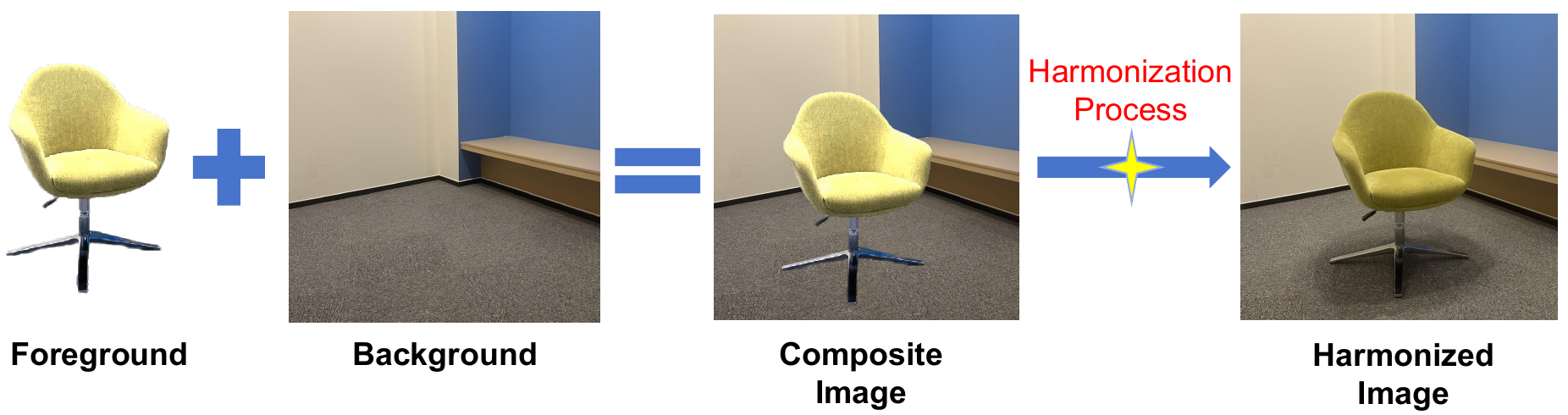} 
\vspace{-15pt}
\caption{Example of a composite image and the harmonization process.}  
\label{harmonization example}  
\end{figure}

\vspace{-5pt}
\label{sec:intro}
Image composition refers to the technique of extracting a foreground object from one image and integrating it into another to create a synthesized composite image \cite{PCT}. However, this process often introduces inconsistencies in statistical attributes such as color and lighting between the foreground and background. These discrepancies can result in composite images that appear unrealistic and visually disharmonious \cite{DoveNet}. Image harmonization algorithms (IHAs) address this issue by adjusting the appearance of the foreground to better match the background, thereby producing more natural and harmonious composite images, as shown in Fig.~\ref{harmonization example}.  

IHAs can be broadly categorized into generative image harmonization algorithms (GIHAs) and non-generative image harmonization algorithms (NGIHAs). Some GIHAs \cite{DIHGAN, PHD} leverage Generative Adversarial Networks (GANs) \cite{GAN}, while others \cite{PHDdiffusion, OS, iclight} employ diffusion models \cite{DM}. More recently, \cite{iclight} has garnered significant attention for its impressive performance in precise illumination manipulation with diffusion model. However, these GIHAs typically rely on self-supervised learning, where ground truths are unavailable \cite{OS}, so it is difficult to evaluate GIHAs. In contrast, NGIHAs are typically trained on datasets generated by altering the style of the foreground in real images \cite{CDT,DoveNet,PCT,DucoNet}. These approaches benefit from the availability of real images as ground truth, enabling quality evaluation through similarity metrics. However, such similarity-based evaluation methods struggle to align with human visual preference. Therefore, there is an urgent need for an effective image quality assessment (IQA) method tailored to the harmonization task, which aligned closely with human perception and applicable to the evaluations of NGIHAs and GIHAs.

\setlength{\tabcolsep}{3pt}
\begin{table*}[t]
\renewcommand{\arraystretch}{0.85}
\centering
\caption{An overview of IHAs selected to construct our HarmonyIQAD.}
\vspace{-5pt}
\label{models}
\begin{tabular}{c|ccccc|cccc}
\toprule
& \multicolumn{5}{c|}{Non-generation IHAs} & \multicolumn{4}{c}{Generation IHAs} \\
Model & L$\&$E\cite{L&E}  &DoveNet\cite{DoveNet}&CDT\cite{CDT}   &PCT\cite{PCT} &DucoNet\cite{DucoNet} &ObjectStitch\cite{OS} &PHD\cite{PHD} &PHDiffusion\cite{PHDdiffusion}&IC-Light\cite{iclight}\\
\midrule
Year &2007.10  & 2020.06&2022.09& 2023.06& 2023.10 &2023.05& 2023.06&2023.08&2024.06\\
Method &Color Statistics  & CNN&U-Net& Transformer& U-Net &Diffusion& GAN&Diffusion&Diffusion\\
Resolution &$256\times256$ &$256\times256$&$2048\times2048$ &$256\times256$ & $1024\times1024$&  $512\times512$ & $512\times512$  &$512\times512$ &$1024\times1024$\\
\bottomrule
\end{tabular}
\vspace{-16.5pt}
\end{table*}
Existing IQA methods can be broudly classified into two categories: full-reference (FR) IQA and no-reference (NR) IQA. FR IQA relies on reference images for comparison, and many classical methods have been proposed in the literature \cite{SSIM,FSIM,NQM,MSSIM,IFC,VIF,GSI,GMSD,VSI}. With the rapid advancements in deep neural networks (DNNs), many studies\cite{LPIPS,STLPIPS,AHIQ} have utilized learning-based methods and achieved outstanding performance. Nevertheless, in practice, FR IQA is not applicable to evaluate the GIHAs due to the lack of reference images. For NR IQA, traditional methods such as \cite{NIQE,BRISQUE,BIQI,DIIVINE,BLII} generally exhibit limited performance. Therefore, learnable NR IQA received more significant attention in recent years, with a variety of works emerging \cite{CNN,Wa,NIMA,DBCNN,Hyper,MANIQA,CLIPIQA,TOPIQ,ARNIQA}, including convolution neural network (CNN) based IQA models and contrastive language-image pretraining (CLIP) based IQA models. Recently, with the rapid development of large language models (LLMs) and vision-language pre-training techniques, some large multimodal models (LMMs) have proven effective in describing image quality \cite{LMMIQA}. However, fine-grained IQA, especially in predicting accurate scores of human visual preference, remains a critical challenge for these LMMs.

In this paper, we address the evaluation of IHAs from the perspective of perceptual quality. Firstly, we construct the first \underline{I}mage \underline{Q}uality \underline{A}ssessment \underline{D}atabase for image \underline{Harmony} evaluation (HarmonyIQAD), including 1350 images, which can be divided into two subsets including a NGIHAs subset and a GIHAs subset. Each subset includes harmonized images generated from representative NGIHAs or GIHAs. Then, we recruit a group of experienced researchers in the field of image-processing to subjectively evaluate the harmonization quality of each image in HarmonyIQAD and obtained MOSs.

Based on HarmonyIQAD, we propose the first a \underline{Harmony} \underline{I}mage \underline{Q}uality \underline{A}ssessment (HarmonyIQA). HarmonyIQA is built upon a LMM \cite{LMM} and incorporates instruction tuning \cite{Ituning} and low-rank adaptation (LoRA) \cite{lora} techniques. A double-stage training is used to achieve the better score regression.
Experimental results demonstrate that HarmonyIQA outperforms all the state-of-the-art FR IQA and NR IQA methods on HarmonyIQAD, and achieves strong performance on other IQA datasets, highlighting its potential as a leading general-purpose IQA method. The key contributions of this work include:
\begin{itemize}
  \item We construct the HarmonyIQAD, the first image harmonization quality database that contains 1350 harmonized images with over 28K subjective quality ratings.
  \item We propose HarmonyIQA, the first evaluator capable of fine-grained image harmonization quality scoring, which is also effective for general image quality assessment.
  \item The extensive experimental results on HarmonyIQAD and other IQA datasets manifest the state-of-art performance of the proposed HarmonyIQA.
\end{itemize}

\vspace{-7pt}
\section{Image Harmonization Quality Database}
\vspace{-2pt}
\begin{figure}[tb]
\centering
\includegraphics[width=0.49\textwidth]{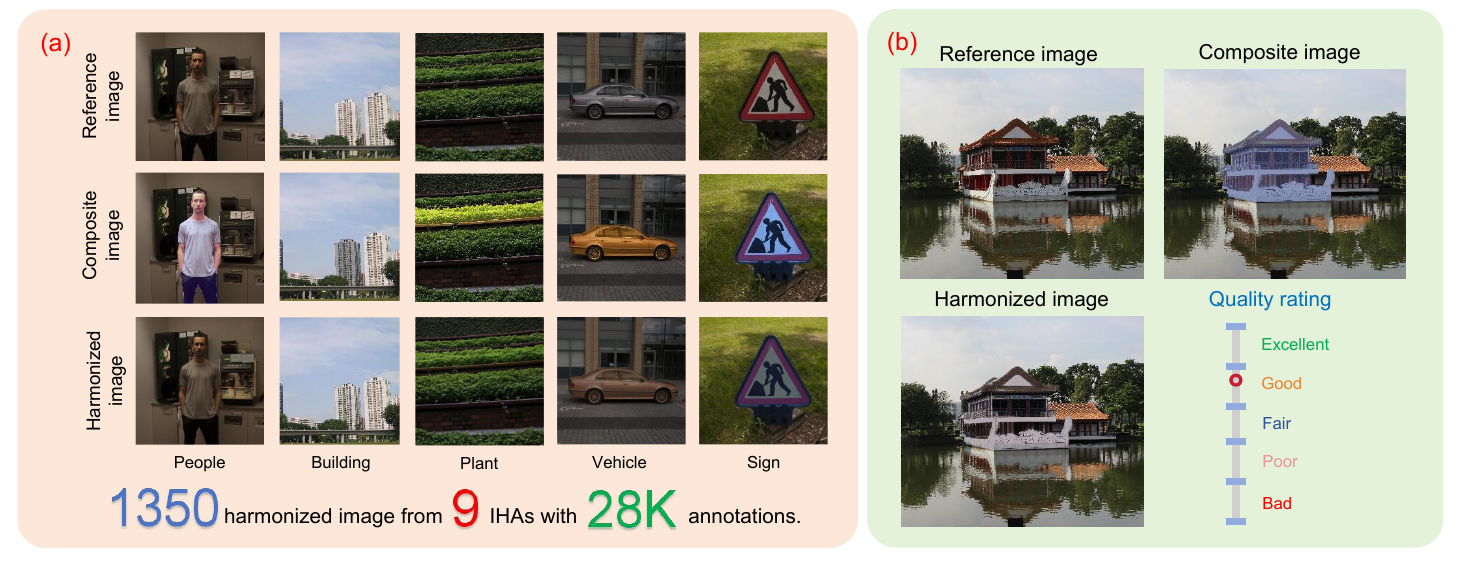}  
\caption{An overview of the content and rating GUI of HarmonyIQAD. (a) Example images from our database, which contains reference images, composite images and harmonization-processed images. (b) The illustration of GUI for subjective rating.}
\label{database}
\end{figure}
In this section, we introduce the proposed HarmonyIQAD. This database comprises 150 composite images, 1,350 harmonization-processed images, 150 reference images, and 28,350 human ratings of image quality. The HarmonyIQAD encompasses a broad range of image content, providing a comprehensive resource for evaluating the quality of image harmonization.
\vspace{-3pt}
\subsection{Image Collection} 
\vspace{-2pt}
The ccHarmony dataset \cite{ccHarmony} forms the basis of our image collection, containing real-world images where foreground objects are placed under varying illumination conditions to create composite images. We selected 150 pairs of composite and reference images, covering diverse foreground objects and color styles. These composite images were processed by 5 NGIHAs, including  L$\&$E \cite{L&E}, DoveNet\cite{DoveNet}, CDT\cite{CDT}, PCT\cite{PCT} and DucoNet\cite{DucoNet}, as well as 4 NGIHAs, namely  ObjectStitch\cite{OS}, PHD\cite{PHD}, PHDdiffusion\cite{PHDdiffusion} and IC-Light\cite{iclight}. The details of these methods are shown in TABLE~\ref{models}. The harmonized images form the GIHAs and NGIHAs subsets, together with the composite and reference images, creating the HarmonyIQAD, as shown in Fig.~\ref{database}(a).

\vspace{-5pt}
\subsection{Subjective Experiment Setup}
\vspace{-2pt}
To evaluate image harmonization quality, we conduct a subjective experiment based on the HarmonyIQAD database. Each time, participants are presented with three images: the harmonized image, composite image, and reference image. They provide a overall quality rating for the harmonized image on a five-point scale, based on three criteria including harmonization effectiveness, content authenticity, and foreground detail preservation. This is because that a good image harmonization algorithm should not only harmonize the composite image, but also maintain the foreground object authenticity and details.

The experiment is conducted using a Python-based GUI displayed on a calibrated LED monitor with a resolution of $3840 \times 2160$, with images shown at their original resolution in random order. A total of $21$ professional annotators, seated $2$ feet from the monitor in a controlled environment, complete the study in $9$ sessions, each under $30$ minutes, to minimize fatigue.
\begin{figure}[tp]
\centering
\includegraphics[width=0.49\textwidth]{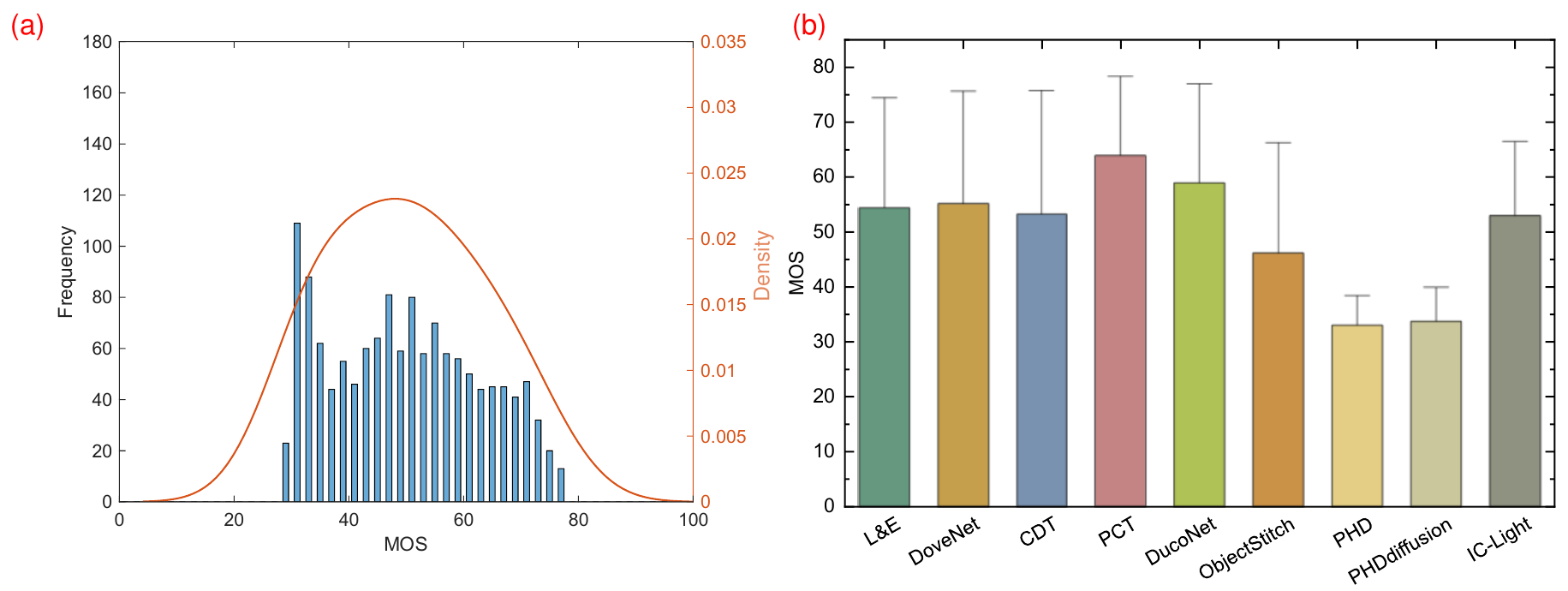}  
\vspace{-20pt}
\caption{(a) MOSs distribution in HarmonyIQAD. (b) Mean and standard deviation of the MOSs for each IHAs.}
\label{dataanalysis}
\end{figure}

\vspace{-5pt}
\subsection{Subjective Data Analysis}
\begin{figure*}[tb]
\centering
\includegraphics[width=0.9\textwidth]{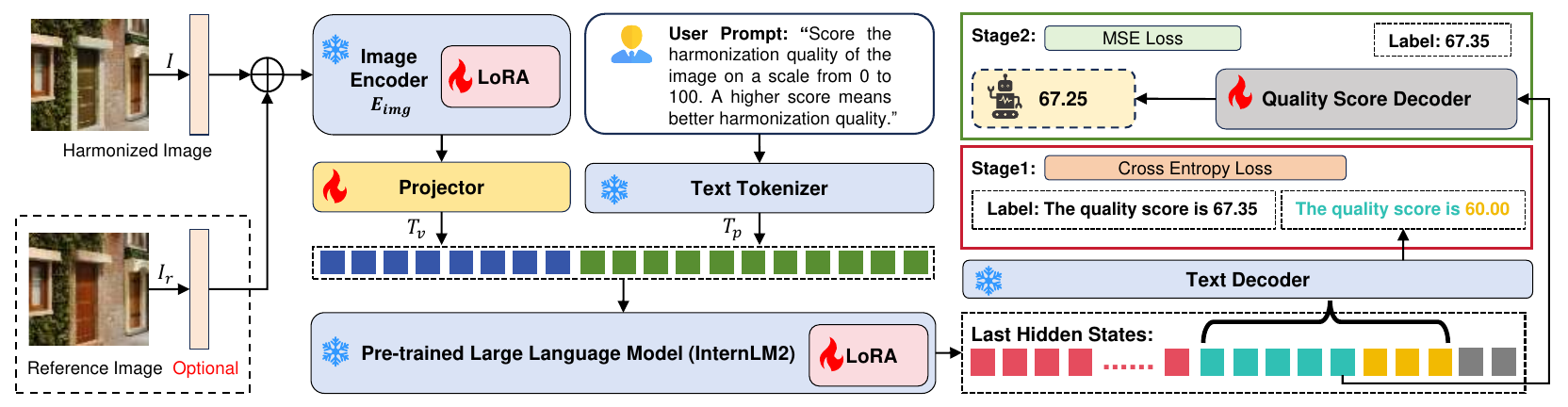}  
\vspace{-10pt}
\caption{The HarmonyIQA model consists of two encoders: a visual encoder for extracting image features and a text encoder for processing user prompt features. These features are aligned by a trainable projector and passed into a pre-trained LLM, from which the last hidden states are selected. In the first training stage, these hidden states are decoded through a text decoder, with the text labels and cross-entropy loss used for training. In the second training stage, the hidden state representing the token just before the score is decoded through a quality score decoder, with the score number and mean squared error loss used for training. LoRA weights are introduced to the vision decoder and LLM to adapt the model for the quality assessment task.}
\vspace{-15pt}
\label{evaluator}
\end{figure*}

We follow the guidelines outlined in \cite{subject} to identify and exclude outliers, as well as to reject subjects who provide unreliable ratings. An individual rating for an image is considered an outlier if it falls outside $2$ standard deviations (if normal) or $\sqrt{20}$
standard deviations (if not normal) from the mean rating of that image. A subject is excluded if over $5\%$ of their ratings are outliers. As a result, no subject was excluded based on this criterion and about $2.72\%$ of the total subjective ratings are removed. The remaining valid ratings are converted into Z-scores, then linearly scaled to the range [0,100]. The final MOS is calculated as followings:
\begin{equation}
    z_{ij} = \frac{r_{ij} - \mu_i}{\sigma_i}, \ z_j = \frac{1}{N_j} \sum_{i=1}^{N_j} z_{ij}, \
    MOS_j = \frac{100(z_j + 3)}{6}
\end{equation}
where where $r_{ij}$ is the raw rating given by the i-th subject to the j-th image, $\mu_i$ is the mean rating and $\sigma_i$ is the standard deviation provided by the i-th subject and $N_j$ is the number of valid ratings for the j-th image.

Based on the MOSs, we further analyze the collected data. Specifically, we visualize the MOS distributions, as shown in Fig.~\ref{dataanalysis}(a). It is clear that the MOS distributions span a wide range. To compare the performance of the Image Harmonization Algorithms (IHAs), we present the mean and standard deviation of the MOSs for the harmonized images produced by each IHAs in Fig.~\ref{dataanalysis}(b). The overall performance of all NGIHAs is similar, and generally higher than that of the GIHAs. Among all the IHAs, PCT achieves the best performance, while PHD has the lowest mean score. Moreover, both NGIHAs and GIHAs have large variability, which indicates that the image content significantly affects IHAs effectiveness.


\section{Image Harmonization Quality Evaluator}
In this section, we present HarmonyIQA, the first image harmonization quality assessment model, designed to predict harmonization quality scores aligned with human conception.

\subsection{Model Design}
\textbf{Overall Architecture.} The overall framework of HarmonyIQA is shown in Fig.~\ref{evaluator}. It takes both harmonized images and user prompts as input and generates a quality score. HarmonyIQA starts by extracting visual and text features from the images and user prompts, respectively. A weight-freezed vision encoder is used to extract image features, which are then projected into a language space via a projector, generating visual tokens $T_v$. For text feature extraction, a tokenizer encodes the user prompt into prompt tokens $T_p$. The concatenated tokens $T_v$ and $T_p$ are fed into a pre-trained LLM. The output consist of last hidden states, which are decoded through a text decoder in the first training stage or a quality score decoder in the second training stage.

\setlength{\tabcolsep}{7pt}
\renewcommand{\arraystretch}{0.85}
\begin{table*}
\fontsize{7.5}{8.5}\selectfont
\caption{Performance of FR IQA methods and the proposed HarmonyIQA (FR) on HarmonyIQAD. The best results are highlighted in \textcolor{mred}{red}, and the second-best results are highlighted in \textcolor{mblue}{blue}.}

\begin{center}
\vspace{-5pt}
\begin{tabular}{lcccccccccccc}
\toprule
\multicolumn{1}{l}{Dataset} & \multicolumn{3}{c}{All HarmonyIQAD} & \multicolumn{3}{c}{NGIHAs Subset} & \multicolumn{3}{c}{GIHAs Subset} \\
\noalign{\vspace{-1.5pt}}
\cmidrule(lr){2-4}
\cmidrule(lr){5-7}
\cmidrule(lr){8-10}
Method/Metrics & SRCC & KRCC & PLCC & SRCC & KRCC & PLCC & SRCC & KRCC & PLCC \\
\noalign{\vspace{-1.5pt}}
\midrule
\noalign{\vspace{-1.5pt}}
MSE & 0.5696 & 0.3976 & 0.4371 & 0.3171 & 0.2204 & 0.3222 & 0.5018 & 0.3485 & 0.3717 \\
PSNR & 0.5696 & 0.3976 & 0.5688 & 0.3171 & 0.2204 & 0.3503 & 0.5018 & 0.3485 & 0.5658 \\
NQM\cite{NQM} & 0.6095 & 0.4252 & 0.6082 & 0.3319 & 0.2261 & 0.3443 & 0.3999 & 0.2707 & 0.4521 \\
MSSIM\cite{MSSIM} & 0.6566 & 0.4682 & 0.6594 & 0.3100 & 0.2175 & 0.3249 & 0.6165 & 0.4339 & 0.6887 \\
SSIM\cite{SSIM} & 0.6153 & 0.4340 & 0.6160 & 0.2514 & 0.1733 & 0.2843 & 0.6378 & 0.4496 & 0.7201 \\
FSIM\cite{FSIM} & 0.6487 & 0.4605 & 0.6536 & 0.2883 & 0.2006 & 0.3051 & 0.6179 & 0.4337 & 0.7035 \\
FSIMC\cite{FSIM} & 0.6655 & 0.4746 & 0.6620 & 0.3533 & 0.2445 & 0.3956 & 0.6160 & 0.4324 & 0.7061 \\
IFC\cite{IFC} & 0.5076 & 0.3438 & 0.3603 & 0.0833 & 0.0548 & 0.1184 & 0.6714 & 0.4700 & 0.7541 \\
VIF\cite{VIF} & 0.5059 & 0.3461 & 0.5492 & 0.1597 & 0.1064 & 0.1525 & 0.6638 & 0.4656 & 0.7632 \\
GSI\cite{GSI} & 0.6155 & 0.4337 & 0.6147 & 0.2797 & 0.1934 & 0.2860 & 0.6065 & 0.4265 & 0.6829 \\
SCSSIM\cite{SCSSIM} & 0.6467 & 0.4588 & 0.6499 & 0.2827 & 0.1958 & 0.3301 & 0.6294 & 0.4449 & 0.7092 \\
GMSD\cite{GMSD} & 0.6817 & 0.4910 & 0.6840 & 0.2968 & 0.2060 & 0.3059 & 0.5979 & 0.4242 & 0.6843 \\
GMSM\cite{GMSD} & 0.6066 & 0.4275 & 0.6061 & 0.2601 & 0.1788 & 0.2570 & 0.6232 & 0.4370 & 0.6933 \\
VSI\cite{VSI} & 0.6705 & 0.4812 & 0.6692 & 0.4892 & 0.3401 & 0.4975 & 0.5521 & 0.3876 & 0.6460 \\
\noalign{\vspace{-1.5pt}}
\midrule
\noalign{\vspace{-1.5pt}}
LPIPS (squeeze) \cite{LPIPS} & 0.5568 & 0.3903 & 0.5262 & 0.4985 & 0.3499 & 0.5002 & 0.5924 & 0.4151 & 0.5177 \\
LPIPS (alex) \cite{LPIPS} & 0.5754 & 0.4051 & 0.6012 & \textcolor{mblue}{0.5199} & \textcolor{mblue}{0.3662} & \textcolor{mblue}{0.5503} & 0.5199 & 0.3662 & 0.5144 \\
LPIPS (vgg) \cite{LPIPS} & 0.5300 & 0.3707 & 0.5611 & 0.3543 & 0.2427 & 0.3368 & 0.6461 & 0.4553 & 0.7154 \\
ST-LPIPS (alex) \cite{STLPIPS} & 0.4135 & 0.2839 & 0.2825 & 0.3637 & 0.2472 & 0.3872 & 0.3467 & 0.2382 & 0.3807 \\
ST-LPIPS (vgg) \cite{STLPIPS} & 0.5104 & 0.3606 & 0.5301 & 0.3570 & 0.2442 & 0.4027 & 0.4443 & 0.3094 & 0.5082 \\
AHIQ \cite{AHIQ} & \textcolor{mblue}{0.7670} & \textcolor{mblue}{0.5677} & \textcolor{mblue}{0.7728} & 0.4213 & 0.2851 & 0.4450 & \textcolor{mblue}{0.7114} & \textcolor{mblue}{0.5175} & \textcolor{mblue}{0.7341} \\
\noalign{\vspace{-1.5pt}}
\midrule
\noalign{\vspace{-1.5pt}}
\rowcolor{gray!20}  
\textbf{HarmonyIQA (FR)} & \textcolor{mred}{0.7988} & \textcolor{mred}{0.6128} & \textcolor{mred}{0.7910} & \textcolor{mred}{0.5965} & \textcolor{mred}{0.4217} & \textcolor{mred}{0.5810} & \textcolor{mred}{0.7146} & \textcolor{mred}{0.5414} & \textcolor{mred}{0.7587} \\
\noalign{\vspace{-2.5pt}}
\bottomrule
\noalign{\vspace{-1.5pt}}
\end{tabular}
\label{performance of FR on our dataset}
\end{center}
\vspace{-18pt}
\end{table*}

\begin{table*}
\caption{Performance of NR IQA methods and the proposed HarmonyIQA (NR) on HarmonyIQAD. The best results are highlighted in \textcolor{mred}{red}, and the second-best results are highlighted in \textcolor{mblue}{blue}.}
\fontsize{7.5}{8.5}\selectfont
\begin{center}
\vspace{-5pt}
\begin{tabular}{lcccc c ccc c ccc}
\toprule
\multicolumn{1}{l}{Dataset} & \multicolumn{3}{c}{All HarmonyIQAD} & \multicolumn{3}{c}{NGIHAs Subset} & \multicolumn{3}{c}{GIHAs Subset} \\
\noalign{\vspace{-1.5pt}}
\cmidrule(lr){2-4}
\cmidrule(lr){5-7}
\cmidrule(lr){8-10}
Method/Metrics & SRCC & KRCC & PLCC & SRCC & KRCC & PLCC & SRCC & KRCC & PLCC \\
\noalign{\vspace{-1.5pt}}
\midrule
BIQI\cite{BIQI} & 0.0835 & 0.0574 & 0.1714 & 0.0385 & 0.0273 & 0.1456 & 0.2027 & 0.1372 & 0.2554 \\
DIIVINE\cite{DIIVINE} & 0.0946 & 0.0620 & 0.1492 & 0.1495 & 0.0991 & 0.1552 & 0.0483 & 0.0305 & 0.1322 \\
BRISQUE\cite{BRISQUE} & 0.1238 & 0.0812 & 0.2287 & 0.0809 & 0.0543 & 0.2166 & 0.2661 & 0.1852 & 0.2259 \\
BLIINDS-II\cite{BLII} & 0.2339 & 0.1531 & 0.3079 & 0.1069 & 0.0736 & 0.1702 & 0.0371 & 0.0241 & 0.0788 \\
NIQE\cite{NIQE} & 0.2527 & 0.1718 & 0.3111 & 0.0959 & 0.0630 & 0.2267 & 0.0649 & 0.0415 & 0.2027 \\
\noalign{\vspace{-1.5pt}}
\midrule
\noalign{\vspace{-1.5pt}}
CNNIQA\cite{CNN} & 0.6215 & 0.4836 & 0.6454 & 0.2286 & 0.2185 & 0.3000 & 0.3952 & 0.3282 & 0.4433 \\
WaDIQaM\cite{Wa} & 0.6204 & 0.4392 & 0.6308 & 0.2695 & 0.1790 & 0.2892 & 0.4227 & 0.2956 & 0.4473 \\
NIMA\cite{NIMA} & 0.5191 & 0.3911 & 0.5055 & 0.2077 & 0.1778 & 0.2721 & 0.3237 & 0.2537 & 0.3962 \\
DBCNN \cite{DBCNN}& 0.6329 & 0.4432 & 0.6664 & 0.1203 & 0.0787 & 0.2011 & 0.5316 & 0.3671 & 0.5827 \\
HyperIQA \cite{Hyper} & 0.6786 & 0.4866 & 0.7041 & 0.2276 & 0.1511 & 0.2410 & 0.5106 & 0.3656 & 0.5265 \\
MANIQA \cite{MANIQA} & \textcolor{mblue}{0.7596} & \textcolor{mblue}{0.5571} & \textcolor{mblue}{0.7583} & \textcolor{mblue}{0.4977} & \textcolor{mblue}{0.3410} & \textcolor{mblue}{0.5265} & 0.6201 & 0.4354 & 0.6630 \\
CLIPIQA \cite{CLIPIQA}& 0.6797 & 0.4822 & 0.6519 & 0.2963 & 0.1935 & 0.1564 & 0.3824 & 0.2571 & 0.3843 \\
TOPIQ \cite{TOPIQ}& 0.7017 & 0.5175 & 0.7047 & 0.2739 & 0.1959 & 0.2767 & \textcolor{mblue}{0.7191} & \textcolor{mblue}{0.5250} & \textcolor{mblue}{0.7583} \\
ARNIQA \cite{ARNIQA}& 0.7046 & 0.5099 & 0.7215 & 0.3036 & 0.2037 & 0.3242 & 0.5797 & 0.4180 & 0.6425 \\
\noalign{\vspace{-1.5pt}}
\midrule
\noalign{\vspace{-1.5pt}}
  \rowcolor{gray!20}  
\textbf{HarmonyIQA (NR)}& \textcolor{mred}{0.7848} & \textcolor{mred}{0.5888} & \textcolor{mred}{0.7650} & \textcolor{mred}{0.7371} & \textcolor{mred}{0.5465} & \textcolor{mred}{0.7261} & \textcolor{mred}{0.8411} & \textcolor{mred}{0.6533} & \textcolor{mred}{0.8318} \\
\noalign{\vspace{-1.5pt}}
\bottomrule
\noalign{\vspace{-1.5pt}}
\end{tabular}
\label{performance of NR on our dataset}
\end{center}
\vspace{-18pt}
\end{table*}

\textbf{Visual Encoding.} The image encoder $E_{img}$ is based on the pre-trained vision transformer, InternViT \cite{internViT}. To align the extracted features with the input space of the LLM, a trainable projector $P_{img}$ with two multi-layer perception (MLP) layers is applied. This projects the image features into a language space, generating the visual feature tokens. For an input image $I$, The process can be formulated as: 
\begin{equation}
    T_v = P_{img}(E_{img}(I))
\end{equation}
where $T_v$ is the visual tokens. HarmonyIQA also supports the input of both harmonized image and reference image $I_r$ to assist the score prediction.

\textbf{Feature Fusion via the LLM.} For a given a user prompt, it is first encoded into text tokens $T_p$ using a text tokenizer. These text tokens $T_p$ are then concatenated with the well-aligned visual tokens $T_v$ to form the input to the LLM. Specifically, the pre-trained InternLM2 \cite{internlm} is used to combine the visual and text tokens for multimodal learning. 

\textbf{Adaptive Decoding}
The last hidden states output by the LLM are decoded by text decoder firstly. Once the model is capable of generating responses in the desired format and content, the hidden state representing the token just before the score is then passed to a quality score decoder. This decoder, consisting of two MLPs, is employed in the second training stage to yield a more precise quality score.
\vspace{-5pt}
\subsection{Fine-tuning Techniques}
\textbf{Instruction Tuning.}
Many recent works have demonstrated the effectiveness of using instruction-tuning strategies for better performance to finish new task \cite{Ituning}. As shown in Fig.~\ref{evaluator}, our user prompt includes clear and explicit problem descriptions, maximizing the model's inference capabilities and enabling HarmonyIQA to accurately respond to specific requirements when performing tasks.

\textbf{LoRA Adaptation.}
To enhance the performance of HarmonyIQA, we employ the LoRA technique \cite{lora} to pre-trained LMM for efficient model adaptation. LoRA models the changes $\Delta W \in R^{d\times k}$ for each layer’s $W \in R^{d\times k}$ as $\Delta W =AB$, where $A \in R^{d\times r}$ and $B\in R^{k\times r}$. The rank $r$ is constrained much smaller than $d$ and $k$ to achieve parameter efficiency. Given the original output
$h = Wx$, the forward pass of LoRA is $h = Wx + \Delta Wx = (W + AB)x$.
With the LoRA, HarmonyIQA can effectively adapt to the image quality scores prediction task.

\textbf{Double-stage Training}
We trained HarmonyIQA in two stages. In the first stage, we used cross-entropy loss, with the label being a sentence that includes the quality score. The goal of this stage is to train the model to generate text in the desired format, along with a roughly accurate quality score. However, relying solely on text training does not yield an accurate score result. Therefore, in the second stage, we use Mean Squared Error (MSE) loss, with the label being the quality score number. The objective of this stage is to refine the ability of HarmonyIQA to produce a more accurate quality score.

\begin{table*}[tp]
\renewcommand{\arraystretch}{0.85}
\centering
\fontsize{7.5}{8.5}\selectfont
\caption{Ablation study of our HarmonyIQAD on HarmonyIQA and its NGIHAs subset and GIHAs subset.}
\vspace{-5pt}
\label{ablation}
\begin{tabular}{ccccccccccc}
\toprule
\noalign{\vspace{-1.5pt}}
\multicolumn{4}{c}{Strategy} & \multicolumn{2}{c}{All HarmonyIQAD} & \multicolumn{2}{c}{NGIHAs Subset} & \multicolumn{2}{c}{GIHAs Subset} \\
\noalign{\vspace{-1.5pt}}
\cmidrule(lr){1-4} \cmidrule(lr){5-6} \cmidrule(lr){7-8} \cmidrule(lr){9-10}
\noalign{\vspace{-1.5pt}}
LoRA(vision) & LoRA(llm) & Projector Training & Quality Score Decoder &  SRCC & PLCC & SRCC  & PLCC & SRCC  & PLCC\\ 
\noalign{\vspace{-1.5pt}}
\midrule
\noalign{\vspace{-1.5pt}}
\checkmark &  &  &  & 0.6972 & 0.6795 & 0.4209 & 0.4141 & 0.5487 & 0.5348 \\
 & \checkmark &  &  &   0.7275  & 0.7107 & 0.4240 & 0.4033 & 0.5231 & 0.5844\\
   \checkmark & \checkmark &  & & 0.7437 & 0.7215 & 0.4328 & 0.4270 & 0.5664 & 0.5925 \\
 \checkmark & \checkmark &  \checkmark && 0.7568 & 0.7393 &0.4374 & 0.4254 &0.6514 &0.6771\\
  \rowcolor{gray!20}  
 \checkmark & \checkmark & \checkmark & \checkmark & \textbf{0.7848} & \textbf{0.7650} & \textbf{0.7371} & \textbf{0.7261} & \textbf{0.8411} & \textbf{0.8318} \\ 
 \noalign{\vspace{-1.5pt}}
\bottomrule
\end{tabular}
\vspace{-16.5pt}
\end{table*}
\setlength{\tabcolsep}{6pt}

\renewcommand{\arraystretch}{0.85}
\begin{table}[t]
\fontsize{7.5}{8.5}\selectfont
\begin{minipage}[t]{0.5\textwidth}  
\centering
\caption{Performance comparison of state-of-the-art NR IQA methods and HarmonyIQA(NR) on CSIQ\cite{CSIQ}, TID2013\cite{TID} and KADID\cite{kadid} databases. The best results are highlighted in \textcolor{mred}{red}, and the second-best results are highlighted in \textcolor{mblue}{blue}.}
\vspace{-5pt}
\label{existing dataset}
\begin{tabular}{lcccccccc}
\toprule
Dataset & \multicolumn{2}{c}{CSIQ} & \multicolumn{2}{c}{TID2013} & \multicolumn{2}{c}{KADID} \\
\noalign{\vspace{-1.5pt}}
\cmidrule(lr){2-3} \cmidrule(lr){4-5} \cmidrule(lr){6-7}
Method/Metrics & SRCC & PLCC & SRCC & PLCC & SRCC & PLCC \\
\noalign{\vspace{-1.5pt}}
\midrule
BRISQUE\cite{BRISQUE} & 0.812 & 0.748 & 0.643 & 0.571 & 0.528 & 0.567 \\
NIQE\cite{NIQE} & 0.627 & 0.721 & 0.315 & 0.393 & 0.374 & 0.428\\
\noalign{\vspace{-1.5pt}}
\midrule
WaDIQaM\cite{Wa} & 0.852 & 0.844 & 0.835 & 0.855 & 0.739 & 0.752 \\
DBCNN\cite{DBCNN} & 0.946 & 0.959 & 0.816 & 0.851 & 0.856 & 0.856 \\
HyperIQA\cite{Hyper} & 0.923 & 0.942 & 0.840 & 0.858 & 0.852 & 0.845 \\
ARNIQA\cite{ARNIQA} & \textcolor{mblue}{0.962} & \textcolor{mblue}{0.973} & \textcolor{mblue}{0.880} & \textcolor{mblue}{0.901} & \textcolor{mblue}{0.908} & \textcolor{mblue}{0.912} \\
\noalign{\vspace{-1.5pt}}
\midrule
\noalign{\vspace{-1.5pt}}
  \rowcolor{gray!20}  
\textbf{HarmonyIQA}& \textcolor{mred}{0.968} & \textcolor{mred}{0.977} & \textcolor{mred}{0.901} & \textcolor{mred}{0.913} & \textcolor{mred}{0.935} & \textcolor{mred}{0.938} \\
\noalign{\vspace{-1.5pt}}
\bottomrule
\end{tabular}
\end{minipage}
\renewcommand{\arraystretch}{0.85}
\begin{minipage}[t]{0.5\textwidth}
\fontsize{7.5}{8.5}\selectfont
\centering
\vspace{-3pt}
\caption{The cross-dataset evaluation results for the SRCC metric using LIVE\cite{LIVE}, CSIQ\cite{CSIQ} and TID2013\cite{TID}. The best scores are highlighted in \textcolor{mred}{red}.}
\vspace{-4pt}
\begin{tabular}{cccccc}
\noalign{\vspace{-1.5pt}}
\toprule
\noalign{\vspace{-1.5pt}}
 \multicolumn{2}{c}{} & \multicolumn{3}{c}{Method} \\
 \noalign{\vspace{-1.5pt}}
\cmidrule(lr){3-5} 

Train on& Test on& HyperIQA\cite{Hyper} & ARNIQA\cite{ARNIQA} & HarmonyIQA \\
\noalign{\vspace{-1.5pt}}
\midrule
\noalign{\vspace{-1.5pt}}
LIVE & CSIQ & 0.744 & \textcolor{mred}{0.904} & 0.838\\
LIVE & TID2013 & 0.541 & 0.697 & \textcolor{mred}{0.784}\\
CSIQ & LIVE & 0.926 & 0.921 & \textcolor{mred}{0.938}\\
CSIQ & TID2013 & 0.541 & 0.721 & \textcolor{mred}{0.787}\\
TID2013 & LIVE& 0.876 & 0.869 &\textcolor{mred}{0.883}\\
TID2013& CSIQ&0.709& \textcolor{mred}{0.866} & 0.804\\
 \noalign{\vspace{-1.5pt}}
\midrule
 \noalign{\vspace{-1.5pt}}
  \rowcolor{gray!20}  
 \multicolumn{2}{c}{Average Performance} &0.723 & 0.830 & \textcolor{mred}{0.839}\\
  \noalign{\vspace{-1.5pt}}
\bottomrule
\label{cross-study}
\end{tabular}
\end{minipage}
\end{table}

\vspace{-8pt}
\section{Experimental Evaluation}
\subsection{Evaluation on HarmonyIQAD}
We first split the HarmonyIQAD dataset into training and test sets with a 4:1 ratio. The harmonization-processed images generated by each IHA are allocated according to this same ratio. Then, 20 FR IQA methods and 14 NR IQA methods are selected as baselines for our HarmonyIQA, totally divided into four groups: traditional FR IQA, leaning-based FR IQA, traditional NR IQA and learning-based NR IQA. Performance is evaluated using three metrics: Spearman rank-order correlation coefficient (SRCC), Kendall rank-order correlation coefficient (KRCC), and Pearson linear correlation coefficient (PLCC). In addition to evaluating on the entire dataset, we also conduct separate evaluations on GIHAs subset and NGIHAs subset. 

The results is shown in TABLE.~\ref{performance of FR on our dataset} and  TABLE.~\ref{performance of NR on our dataset}. We find that HarmonyIQA outperforms all other IQA methods across all metrics on both the whole and subsets of HarmonyIQAD.

\vspace{-9pt}
\subsection{Ablation Study}
\vspace{-2pt}
We conduct ablation studies on HarmonyIQAD and its two subsets to evaluate the effectiveness of the core components of HarmonyIQA. The results are presented in TABLE~\ref{ablation}. 

\textbf{Effectiveness of LoRA Adaptation.}
We first demonstrate the effectiveness of LoRA adaptation, as shown in rows 1 to 3 of TABLE~\ref{ablation}. The results indicate that applying LoRA to both the vision model and the LLM yields the best performance for the our HarmonyIQA.

\textbf{Effectiveness of Training Projector Weights.}
The projector in HarmonyIQA is responsible for projecting vision features into the language space. As shown in row 4 of TABLE~\ref{ablation}, training the projector weights results in improved performance.

\textbf{Effectiveness of Quality Score Decoder.}
HarmonyIQA undergoes two stages of training. In the second training stage, a quality score decoder is employed for more accurate quality score regression. Row 5 of TABLE~\ref{ablation} demonstrates that adding the quality score decoder significantly improves performance.


\vspace{-9pt}
\subsection{Evaluation on Existing IQA datasets}
\vspace{-3pt}
We further evaluate the performance of the proposed HarmonyIQA on three additional IQA benchmark datasets: CSIQ\cite{CSIQ}, TID2013\cite{TID} and KADID\cite{kadid}. For these three datasets, we follow the same principles of splitting the data into training and test sets with a 4:1 ratio. TABLE.~\ref{existing dataset} presents that HarmonyIQA achieves state-of-the-art performance across these datasets.
\vspace{-8pt}
\subsection{Cross-dataset Evaluation}
\vspace{-3pt}
We also conduct cross-database evaluations on three IQA datasets including LIVE\cite{LIVE}, CSIQ\cite{CSIQ} and TID2013\cite{TID}. The results shown in TABLE~\ref{cross-study} demonstrate that, compared to two other state-of-the-art IQA methods \cite{Hyper, ARNIQA}, HarmonyIQA demonstrates superior performance. In particular, while ARNIQA employs self-supervised learning and linear regression, HarmonyIQA exhibits better overall generalization ability over this method.


\section{Conclusion}
In this paper, we present HarmonyIQAD, the first comprehensive image harmonization quality database, which contains 1,350 harmonized images with subjective quality scores. Based on the dataset, we propose HarmonyIQA, which leverages advanced visual encoding techniques, instruction tuning and LoRA tuning, to perform image quality predicting. Extensive experiments on HarmonyIQAD and other widely used IQA datasets demonstrate the excellent performance of HarmonyIQA, highlighting its potential for general image quality assessment application.

\bibliographystyle{IEEEbib}
\vspace{-10pt}
\bibliography{main}

\end{document}